\documentclass{article}
\usepackage{spconf,amsmath,graphicx}
\usepackage{booktabs}
\usepackage{xcolor}
\usepackage{url}
\usepackage{enumitem}
\setlist{nosep, leftmargin=14pt}

\newif\ifshowtodos
\showtodostrue

\title{TAN: A Size- and Form-Adaptive Masked-Convolution Encoder for Tissue Graphs}
\name{Jan Herschel$^{\star}$ \qquad Anselm von Gladiss$^{\star \dagger}$}
\address{$^{\star}$ Institute for Computer Science, University of Koblenz, Germany \\
         $^{\dagger}$ MTI Mittelrhein, University of Koblenz, Germany}
\begin{document}
\maketitle

\begin{abstract}
Tissue graphs represent histology images as graphs over tissue-conforming superpixels, which keeps homogeneous tissue regions intact instead of cutting them into rectangular patches.
To embed superpixels, current encoders crop them to bounding boxes and then pad, resize, or mask them, so background or fill values can leak into the node features.
This matters in practice, and especially in medical imaging, where a prediction should depend on the tissue and not on an arbitrary preprocessing choice: on BACH, changing only the fill of the masked region shifts the accuracy of frozen encoders by up to 6 percentage points, larger than the differences among the four best encoders under mean fill, and the preferred fill differs between datasets, so it cannot be fixed once and for all.
We present TAN (Tissue-Aware Network), a compact convolutional encoder (0.41M parameters) that processes each superpixel at its native shape and extent and excludes background from every operation, which makes its embeddings exactly invariant to the fill by construction.
Under identical masks and an identical graph classifier, TAN's accuracy on BACH is not significantly different from that of the 303M-parameter foundation model UNI (81.1\% vs.\ 80.7--82.6\%, depending on the fill), with ${\approx}740\times$ fewer parameters, ${\approx}1{,}190\times$ fewer FLOPs, and no pretraining.
On BRACS, TAN is close to a ResNet18 trained under the same protocol and trails the histopathology foundation models.
An ablation with TAN's architecture but standard processing of fill-padded crops shows that masking keeps accuracy similar (nominally higher on BACH, lower on BRACS, within noise) at less than half the FLOPs.
\end{abstract}

\section{Introduction}
\label{sec:intro}

Whole Slide Images (WSIs) span gigapixel dimensions and are therefore usually cut into rectangular patches and classified with Multiple Instance Learning (MIL) \cite{Wu2026,10654803,Xu2025}.
MIL treats the patches as independent instances and ignores where they lie, so it cannot represent how tissue regions are arranged and connected \cite{Wu2026}. 
Tissue graphs address this by encoding local microenvironments and global tissue architecture \cite{pmlr-v156-jaume21a}: nodes, commonly superpixels, represent tissue regions, and edges their spatial relationships.
This work targets node-feature encoding.

Current formulations struggle with non-rectangular nodes.
Node features come from handcrafted color and texture descriptors \cite{10.1007/978-3-030-60365-6_20}, which lack the expressive capacity of deep representations, or from deep-learning backbones restricted to rectangular grids, e.g., applied to fixed-size patches centered on each region \cite{PATI2022102264}.
Forcing an irregular entity into such a grid requires cropping, resizing, padding, or masking a rectangular crop, which can introduce artifacts, distort scale, and mix surrounding content into the node features; for standard convolutions over masked regions filled with substitute values, artifacts are well documented in inpainting \cite{Liu_2018_ECCV}.

We address this with \textbf{TAN} (Tissue-Aware Network), a compact encoder built on a strict masked-convolution rule that consumes each superpixel at its native shape and extent, with background excluded from every operation.
To separate the effect of masked processing from those of architecture and pretraining, we compare it on BACH \cite{ARESTA2019122} and BRACS \cite{Brancati2022} with an unmasked twin of identical architecture, a ResNet18 trained from scratch, and frozen ImageNet, CTransPath and UNI encoders.

\section{Related Work}
\label{sec:related}
\emph{Node features of tissue graphs.}
Tissue graphs use superpixels or other tissue components as nodes \cite{pmlr-v156-jaume21a}; node features are handcrafted color and texture descriptors \cite{10.1007/978-3-030-60365-6_20} or deep features of fixed-size patches centered on each region \cite{PATI2022102264}, which do not follow the region's shape.

\emph{Pathology foundation models.}
UNI \cite{Chen2024} and CTransPath \cite{Wang2022} are pretrained on large histopathology corpora and serve as patch-level feature extractors in MIL pipelines \cite{Xu2025}; they take rectangular inputs, so an irregular superpixel again needs a fill for its surroundings.

\emph{Masked convolutions.}
Partial convolution \cite{Liu_2018_ECCV} masks invalid pixels and renormalizes by the number of valid ones, but already produces an output when a window contains a single valid pixel, so the valid mask grows outward with depth; gated convolution \cite{Yu_2019_ICCV} learns a soft gate, so excluding masked input is learned rather than guaranteed; submanifold sparse convolution \cite{8579059} keeps the active sites fixed and restricts normalization and activations to them, but treats inactive inputs as zero, an implicit zero fill at region boundaries.
None of them restricts outputs to windows that lie entirely inside the region.

\section{Materials and Methods}
\label{sec:method}

\subsection{Datasets}
\label{sec:datasets}
\textbf{BACH} \cite{ARESTA2019122}: 400 H\&E images, 4 balanced classes (accuracy reported), native resolution (0.42\,$\mu$m/pixel), SLIC target of 800 superpixels per image (625 on average); 5-fold stratified cross-validation repeated 3 times (15 folds), identical across encoders.

\textbf{BRACS} \cite{Brancati2022}: 7-class breast lesion subtyping on the official train/val/test split. RoIs are scanned at 0.25\,$\mu$m/pixel and vary in size, so each image is center-cropped to $1000{\times}1000$\,px (smaller images, 31\%, are excluded, leaving 2{,}584/221/321 of the official 3{,}657/312/570 train/val/test images) and segmented with a target of 250 superpixels (197 on average), matching BACH's superpixel size in pixels (median 4{,}645 vs.\ 4{,}617) but not in tissue area, which is about a third (290 vs.\ 810\,$\mu$m$^2$).
With a fixed split, TAN, TAN-unmasked, ResNet18-scratch and each GAT are trained with 5 seeds and reported as mean $\pm$ std over seeds.
After the filter, the training split holds 213 (UDH) to 591 (PB) images per class and the test split 25 (FEA) to 64 (IC), so macro-F1 is reported alongside accuracy; the filter is class-skewed, keeping 90\% of IC but only 38\% of FEA, 55\% of UDH and 59\% of ADH images.
\subsection{Superpixel Graph Construction}
\label{sec:graph}
Images are segmented with SLIC \cite{6205760}; the superpixels, without further merging, are the nodes of the tissue graph, and two nodes are connected by an undirected edge if their regions share a pixel border.
Every encoder sees only its own superpixel's masked content, so node embeddings are context-free: neighborhood information enters only through the graph classifier.
For encoders with standard convolution, pixels outside the mask are replaced by the superpixel's mean color (\emph{mean fill}) or by constant white (\emph{white fill}); results are reported for both.
TAN receives the raw crop plus the mask and needs no fill value.

\subsection{TAN: Tissue-Aware Network}
\label{sec:tan}

\begin{figure}[t]
\centering
\includegraphics[width=0.5\linewidth]{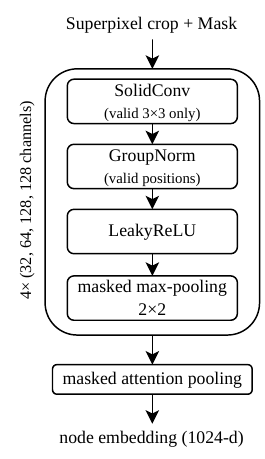}
\caption{TAN encoder: four stages of masked convolution (SolidConv), mask-aware GroupNorm, LeakyReLU, and masked $2{\times}2$ max-pooling (32, 64, 128, 128 channels), then masked attention pooling to a 1024-d node embedding. The mask is downsampled alongside the features; invalid positions are ignored in every operation.}
\label{fig:tan_architecture}
\end{figure}
\begin{figure*}[t]
\centering
\includegraphics[width=\textwidth]{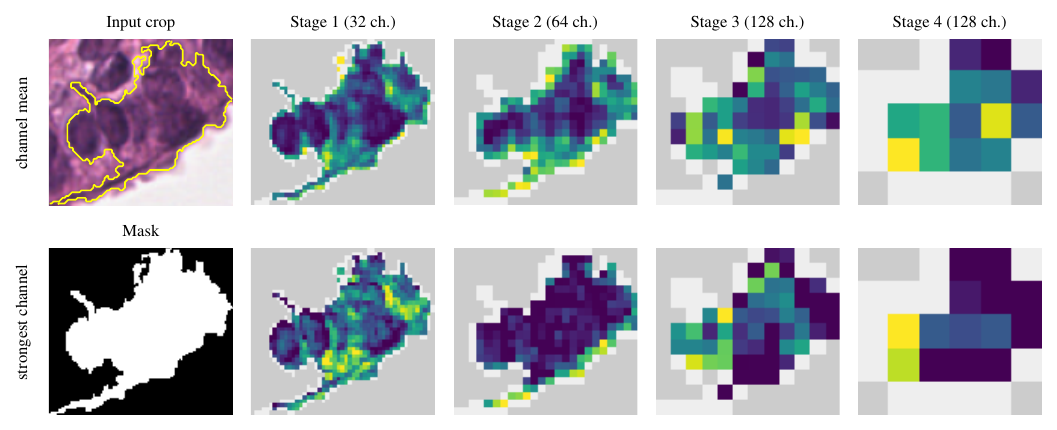}
\caption{A superpixel from a BRACS test image passing through a trained TAN. Top row: input crop with the superpixel outline and, for stages 1--4, the channel mean of each stage's output. Bottom row: the mask and the strongest channel of each stage, i.e., the channel with the largest spatial variance. Dark gray: outside the mask; light gray: inside the mask but without a fully valid $3{\times}3$ neighborhood (bias only). Panels are scaled independently.}
\label{fig:tan_stages}
\end{figure*}

TAN is a 4-stage encoder (Fig.~\ref{fig:tan_architecture}). Each stage applies a masked convolution (\emph{SolidConv}: a pixel's output is computed only if its full $3\times3$ neighborhood lies within the current mask), mask-aware GroupNorm (statistics over valid positions), LeakyReLU, and masked $2\times2$ max-pooling (invalid positions cannot win), with 32, 64, 128 and 128 channels.
An 8-head attention pooling, whose softmax excludes invalid positions, yields a 1024-d embedding that feeds a linear classifier head.
TAN passes the superpixel mask on unchanged, only downsampling it by pooling between stages, so the valid region keeps the superpixel's shape (up to pooling resolution), and a position receives a data-dependent output only if its entire $3\times3$ neighborhood lies within the mask, so boundary outputs never mix in implicit zeros.
The outermost ring of valid positions at each stage therefore carries only the layer's normalized bias term and no local image content (light gray in Fig.~\ref{fig:tan_stages}).
Superpixels without any fully valid $3\times3$ neighborhood at some stage carry no image content: 0.001\% (3 of 249{,}800) on BACH, 0.003\% (16 of 616{,}769) on BRACS.
Crops of similar area are batched and zero-padded to the batch maximum; the padded values lie outside the mask and cannot contaminate any operation (embeddings are identical to float precision when the padding preserves the pooling alignment).
TAN is not pretrained externally: each superpixel is trained to predict its image's class, a weak, noisy but dense signal.
On BACH, TAN is retrained for every cross-validation fold (15\% of the fold's training images held out for early stopping); on BRACS it is trained on the filtered official training split (2{,}584 images) with the filtered official validation split (221 images) for early stopping, once per seed. TAN, TAN-unmasked and ResNet18-scratch share the augmentation function (flips, 90-degree rotations, intensity jitter) and the epoch limit, patience, learning rate and weight decay; batch sizes differ (Section~\ref{sec:encoders}).

\subsection{Comparison Encoders}
\label{sec:encoders}
\textbf{TAN-unmasked} isolates masked processing: it has TAN's architecture, parameters (408{,}580), augmentation, size-sorted batching and training settings (batch size 16), but the pixels outside the mask are replaced by the fill (batches are padded with each crop's own fill color) and processed like any other pixel by standard convolution, GroupNorm, max-pooling and attention over all positions; it is evaluated on both datasets.
\textbf{ResNet18-scratch} is a standard randomly initialized ResNet18 \cite{He2016} (11{,}178{,}564 parameters) trained with the same protocol as TAN (same superpixel-level image-label supervision, per-fold retraining, epochs, patience, learning rate and weight decay; batch size 32 vs.\ 16) on a masked $128\times128$ crop: an architecture control, evaluated on both datasets.
\textbf{ResNet18-ImageNet}, \textbf{CTransPath} \cite{Wang2022} and \textbf{UNI} \cite{Chen2024} are frozen encoders pretrained on ImageNet, TCGA+PAIP, and a large, mostly private histopathology corpus. Masked crops are extracted at each encoder's input resolution ($128\times128$; $224\times224$ for CTransPath and UNI), never resized, and normalized with ImageNet channel statistics.

\subsection{Downstream Classifier and Protocol}
\label{sec:downstream}
A 3-layer GAT \cite{VelickovicCCRLB18} (4 attention heads per layer) with mean/max pooling classifies each graph from its node embeddings.
As is usual for UNI and CTransPath, the frozen backbones (like ResNet18-ImageNet) are not adapted; only the GAT is trained on their embeddings.
A Random Forest on pooled (mean/std/min/max/median) per-image statistics of the node features is a non-graph reference.

\subsection{Statistical Methodology}
\label{sec:stats}
On BACH, folds of repeated cross-validation share training data and, across repeats, test data, so naive paired tests understate the variance \cite{Dietterich1998,NIPS1999_7d12b66d}.
We use the Nadeau--Bengio corrected resampled $t$-test \cite{NIPS1999_7d12b66d}, applied to repeated $k$-fold cross-validation as in \cite{Bouckaert2004}, which scales the variance by $(1/J + n_{\text{test}}/n_{\text{train}})$ with $J=15$ the total number of train/test splits (5 folds $\times$ 3 repeats), with Holm--Bonferroni correction over the five comparisons against TAN (per fill condition), and call differences significant at corrected $p<0.05$.
BRACS has a single split, so we report seed means and standard deviations without tests.

\section{Results}
\label{sec:experiments}

\subsection{Results on BACH}
\label{sec:bach}

\begin{table}[t]
\centering
\footnotesize
\setlength{\tabcolsep}{3pt}
\caption{BACH, 15-fold repeated CV, accuracy (mean $\pm$ std, \%). GFLOPs: forward pass per superpixel (1 multiply-accumulate, MAC, = 2 FLOPs; mean of 300 sampled superpixels; PyTorch FLOP counter; excludes training and the GAT). TAN's parameters include its classifier head (0.28M without). RF baseline: $62.5 \pm 5.2$.}
\label{tab:bach}
\begin{tabular}{lrrcc}
\toprule
Encoder & Params & GFLOPs & Mean fill & White fill \\
\midrule
TAN                & 0.41M   & 0.10   & \multicolumn{2}{c}{$81.1 \pm 4.5$} \\
TAN-unmasked       & 0.41M   & 0.22   & $80.9 \pm 3.9$ & $79.3 \pm 4.2$ \\
UNI                & 303.4M  & 119.3  & $80.7 \pm 3.7$ & $82.6 \pm 4.4$ \\
ResNet18-scratch   & 11.2M   & 1.18   & $76.8 \pm 3.7$ & $75.6 \pm 6.2$ \\
CTransPath         & 27.5M   & 8.99   & $71.2 \pm 6.2$ & $75.6 \pm 5.5$ \\
ResNet18-ImageNet  & 11.2M   & 1.18   & $65.2 \pm 5.2$ & $71.3 \pm 4.8$ \\
\bottomrule
\end{tabular}
\end{table}

Table~\ref{tab:bach} gives accuracy for both fills; differences between accuracies are in percentage points (pp).
TAN is numerically first under mean fill and second to UNI under white fill, and is not significantly different from UNI or from TAN-unmasked in either case (corrected $p=1.0$).
It is 4.3\,pp above ResNet18-scratch under mean fill (better in 12 of 15 folds), but not significantly after correction ($p=0.49$; $0.72$ under white fill), and a non-significant difference does not establish equivalence: with 15 folds the test has limited power.
Under mean fill TAN is 9.9\,pp above CTransPath (14 of 15 folds), which does not survive correction ($p=0.065$; uncorrected $0.016$), and clearly above ResNet18-ImageNet ($p=0.001$); under white fill neither is significant ($p=0.62$, $0.08$).

The ablation isolates masked processing: TAN-unmasked matches TAN under mean fill ($80.9$ vs.\ $81.1$; TAN better in 9 of 15 folds) and is 1.8\,pp lower under white fill ($79.3$; corrected $p=1.0$, uncorrected $0.61$).
Masking thus does not improve accuracy on BACH. Its measurable effects are compute (0.10 vs.\ 0.22 GFLOPs per superpixel, ${\approx}2.2\times$) and fill dependence: TAN-unmasked changes by 1.6\,pp between the two fills, whereas a trained TAN's embeddings were bit-identical under four backgrounds (original, white, mean, random noise) on 100 random BRACS superpixels, while a frozen ResNet18's differed by a median of 64\% of their norm between white and mean fill.

TAN has ${\approx}27\times$ and ${\approx}740\times$ fewer parameters than ResNet18-scratch and UNI and needs on average 0.10 GFLOPs per superpixel (std 0.05), $\approx$1{,}190$\times$ fewer than UNI and ${\approx}11.8\times$ fewer than ResNet18 at $128\times128$, which process a full window per superpixel, mostly fill.
The background fill changes the frozen encoders by up to 6\,pp (white fill: ResNet18-ImageNet $+6.1$, CTransPath $+4.4$, UNI $+1.9$), more than the differences among the four best encoders under mean fill (at most 4.3\,pp); the from-scratch networks change less ($-1.2$ ResNet18-scratch, $-1.6$ TAN-unmasked; within fold-level std).

\subsection{Results on BRACS}
\label{sec:bracs}
\begin{table}[t]
\centering
\footnotesize
\setlength{\tabcolsep}{2.5pt}
\caption{BRACS, official split, 5 seeds (mean $\pm$ population std over seeds, \%). RF baseline: 32.0 / 26.1.}
\label{tab:bracs}
\begin{tabular}{lcccc}
\toprule
 & \multicolumn{2}{c}{Accuracy} & \multicolumn{2}{c}{Macro-F1} \\
Encoder & Mean & White & Mean & White \\
\midrule
TAN               & \multicolumn{2}{c}{$45.8 \pm 3.5$} & \multicolumn{2}{c}{$42.0 \pm 3.0$} \\
TAN-unmasked      & $49.7 \pm 4.1$ & $46.4 \pm 3.0$ & $46.4 \pm 3.5$ & $43.8 \pm 2.3$ \\
UNI               & $53.3 \pm 2.1$ & $49.4 \pm 2.4$ & $50.7 \pm 2.3$ & $46.4 \pm 2.6$ \\
ResNet18-scratch  & $47.0 \pm 2.1$ & $45.2 \pm 2.4$ & $43.6 \pm 2.3$ & $42.1 \pm 3.8$ \\
CTransPath        & $50.5 \pm 3.7$ & $47.5 \pm 2.2$ & $47.7 \pm 3.0$ & $45.3 \pm 1.4$ \\
ResNet18-ImageNet & $46.1 \pm 1.5$ & $42.9 \pm 2.1$ & $42.6 \pm 1.9$ & $39.6 \pm 1.7$ \\
\bottomrule
\end{tabular}
\end{table}
Table~\ref{tab:bracs} gives accuracy and macro-F1. All encoders exceed the Random Forest baseline.
TAN's accuracy is close to that of ResNet18 trained from scratch ($47.0$ and $45.2$ under mean and white fill) and of ImageNet-pretrained ResNet18 ($46.1$ and $42.9$), but below UNI and CTransPath: by 7.5 and 4.7\,pp under mean fill, 3.6 and 1.7\,pp under white fill. As on BACH, the ResNet18 trained with the same protocol shows no clear accuracy advantage over TAN.
TAN-unmasked reaches $49.7$ and $46.4$ under mean and white fill, $3.9$ and $0.6$\,pp above TAN; these differences are within noise (seed std 3--4\,pp), so masking brings no accuracy benefit here either.
Mean fill is better than white fill for TAN-unmasked (3.3\,pp, as on BACH) and for all three frozen encoders (3.0--3.9\,pp), for which BACH showed the opposite, so neither fill is better in general.

\section{Limitations}
\label{sec:limitations}
\emph{Task setting.}
Superpixels have no labels of their own (TAN is trained with the class of its image, a multiple-instance setting), and the tissue structures they contain are not annotated; together with context-free embeddings, this weak supervision may explain the low absolute accuracies, which are not comparable to image-level results (all experiments use RoIs, not complete WSIs).

\emph{Generality.}
BRACS has one split and a class-skewed filter, and TAN's embeddings depend on batch padding through floor pooling (padding to multiples of 16 should remove this).
The superpixel size was matched in pixels, not in tissue area (Section~\ref{sec:datasets}), and was not tuned; this difference in scale could contribute to the different results of TAN relative to the foundation models on the two datasets, and further experiments with other datasets and resolutions and a study of the superpixel size are needed to evaluate this and to find the best superpixel size.

\section{Conclusion}
\label{sec:conclusion}
We introduced TAN, a 0.41M-parameter encoder that keeps tissue regions intact and excludes background from every operation, so its embeddings are exactly invariant to the fill, which matters in medical imaging.
On BACH its accuracy is not significantly different from the 303M-parameter UNI, with ${\approx}740\times$ fewer parameters, ${\approx}1{,}190\times$ fewer FLOPs and no pretraining; on BRACS it is close to ResNet18 but behind UNI and CTransPath.
An unmasked twin reaches similar accuracy (nominally lower on BACH, higher on BRACS) at more than twice the FLOPs.
Whether pretraining or a better superpixel size closes the BRACS gap remains open.

Code: \url{https://github.com/jherschel94/tan}

\section*{Compliance with Ethical Standards}
This research study was conducted retrospectively using human subject data made available in open access by the BACH \cite{ARESTA2019122} and BRACS \cite{Brancati2022} dataset organizers, who report that the images were de-identified. No ethical approval was sought, since only publicly available, de-identified data were analyzed and the datasets were used in accordance with their terms of use. 

\section*{Acknowledgments}
Claude Sonnet 5.5 (Anthropic) was used as an assistant under the authors' direction. \emph{Code:} Claude implemented most of the pipeline (graph construction, encoders, training and evaluation, statistical tests, FLOP counting, figure scripts) from the authors' specifications; the authors ran, checked and corrected it. \emph{Text:} Claude wrote and revised wording for the Abstract and Sections 1--6 and the \LaTeX{} source from the authors' instructions, results and edits; the authors specified the content, decided what to claim, and edited the text. The authors designed the study, ran all experiments, checked the reported numbers against the raw results, and take full responsibility for the content.
No funding was received for conducting this study. The authors have no relevant financial or non-financial interests to disclose.

\bibliographystyle{IEEEbib}
\bibliography{strings,refs}

\begin{thebibliography}{10}

\bibitem{Wu2026}
Jiyang Wu and Dongxun Jiang,
\newblock ``Pathology-prior driven substructure-aware graph neural network for
  whole slide image classification,''
\newblock {\em Sci. Rep.}, vol. 16, 2026.

\bibitem{10654803}
Jiangbo Shi et~al.,
\newblock ``{ViLa-MIL}: Dual-scale vision-language multiple instance learning
  for whole slide image classification,''
\newblock in {\em Proc. IEEE/CVF CVPR}, 2024, pp. 11248--11258.

\bibitem{Xu2025}
Hongming Xu et~al.,
\newblock ``When multiple instance learning meets foundation models: Advancing
  histological whole slide image analysis,''
\newblock {\em Med. Image Anal.}, vol. 101, pp. 103456, 2025.

\bibitem{pmlr-v156-jaume21a}
Guillaume Jaume et~al.,
\newblock ``{HistoCartography}: A toolkit for graph analytics in digital
  pathology,''
\newblock in {\em Proc. MICCAI Workshop Comput. Pathol., PMLR 156}, 2021, pp.
  117--128.

\bibitem{10.1007/978-3-030-60365-6_20}
Pushpak Pati et~al.,
\newblock ``{HACT-Net}: A hierarchical cell-to-tissue graph neural network for
  histopathological image classification,''
\newblock in {\em Proc. MICCAI Workshops UNSURE and GRAIL}, 2020, pp. 208--219.

\bibitem{PATI2022102264}
Pushpak Pati et~al.,
\newblock ``Hierarchical graph representations in digital pathology,''
\newblock {\em Med. Image Anal.}, vol. 75, pp. 102264, 2022.

\bibitem{Liu_2018_ECCV}
Guilin Liu et~al.,
\newblock ``Image inpainting for irregular holes using partial convolutions,''
\newblock in {\em Proc. ECCV}, 2018.

\bibitem{ARESTA2019122}
Guilherme Aresta et~al.,
\newblock ``{BACH}: Grand challenge on breast cancer histology images,''
\newblock {\em Med. Image Anal.}, vol. 56, pp. 122--139, 2019.

\bibitem{Brancati2022}
Nadia Brancati et~al.,
\newblock ``{BRACS}: A dataset for {BReAst} carcinoma subtyping in {H\&E}
  histology images,''
\newblock {\em Database}, vol. 2022, pp. baac093, 2022.

\bibitem{Chen2024}
Richard~J. Chen et~al.,
\newblock ``Towards a general-purpose foundation model for computational
  pathology,''
\newblock {\em Nat. Med.}, vol. 30, no. 3, pp. 850--862, 2024.

\bibitem{Wang2022}
Xiyue Wang et~al.,
\newblock ``Transformer-based unsupervised contrastive learning for
  histopathological image classification,''
\newblock {\em Med. Image Anal.}, vol. 81, pp. 102559, 2022.

\bibitem{Yu_2019_ICCV}
Jiahui Yu et~al.,
\newblock ``Free-form image inpainting with gated convolution,''
\newblock in {\em Proc. IEEE/CVF ICCV}, 2019, pp. 4470--4479.

\bibitem{8579059}
Benjamin Graham, Martin Engelcke, and Laurens {van der Maaten},
\newblock ``3{D} semantic segmentation with submanifold sparse convolutional
  networks,''
\newblock in {\em Proc. IEEE/CVF CVPR}, 2018, pp. 9224--9232.

\bibitem{6205760}
Radhakrishna Achanta et~al.,
\newblock ``{SLIC} superpixels compared to state-of-the-art superpixel
  methods,''
\newblock {\em IEEE Trans. Pattern Anal. Mach. Intell.}, vol. 34, no. 11, pp.
  2274--2282, 2012.

\bibitem{He2016}
Kaiming He, Xiangyu Zhang, Shaoqing Ren, and Jian Sun,
\newblock ``Deep residual learning for image recognition,''
\newblock in {\em Proc. IEEE CVPR}, 2016, pp. 770--778.

\bibitem{VelickovicCCRLB18}
Petar Veli{\v{c}}kovi{\'c} et~al.,
\newblock ``Graph attention networks,''
\newblock in {\em Proc. ICLR}, 2018.

\bibitem{Dietterich1998}
Thomas~G. Dietterich,
\newblock ``Approximate statistical tests for comparing supervised
  classification learning algorithms,''
\newblock {\em Neural Comput.}, vol. 10, no. 7, pp. 1895--1923, 1998.

\bibitem{NIPS1999_7d12b66d}
Claude Nadeau and Yoshua Bengio,
\newblock ``Inference for the generalization error,''
\newblock in {\em Proc. NeurIPS}, 1999, vol.~12.

\bibitem{Bouckaert2004}
Remco~R. Bouckaert and Eibe Frank,
\newblock ``Evaluating the replicability of significance tests for comparing
  learning algorithms,''
\newblock in {\em Proc. PAKDD}, 2004, pp. 3--12.

\end{thebibliography}

\end{document}